\documentclass[
twocolumn,
]{ceurart}
\usepackage{appendix}
\usepackage{subcaption}
\usepackage{hyperref}
\usepackage{listings}
\begin{document}

\copyrightyear{2022}
\copyrightclause{Copyright for this paper by its authors.
  Use permitted under Creative Commons License Attribution 4.0
  International (CC BY 4.0).}

\conference{Work in Progress}

\title{Evaluating Environments Using Exploratory Agents}


\author[1]{Bobby Khaleque}
[%
orcid=0009-0000-3039-4694,
email=b.d.a.khaleque@qmul.ac.uk,
url=https://yamadharma.github.io/,
]
\cormark[1]
\fnmark[1]
\address[1]{Queen Mary University of London}
\address[2]{Kings College London}

\author[2]{Mike Cook}
[
orcid=0000-0001-5898-9884,
email=mike@possibilityspace.org,
url=https://kmitd.github.io/ilaria/,
]
\fnmark[1]

\author[3]{Jeremy Gow}[
orcid=0009-0004-2768-6898,
email=jeremy.gow@qmul.ac.uk,
url=http://conceptbase.sourceforge.net/mjf/,
]
\address[3]{Queen Mary University of London}



\begin{abstract}
Exploration is a key part of many video games. We investigate the using an exploratory agent to provide feedback on the design of procedurally generated game levels, 5 engaging levels and 5 unengaging levels. We expand upon a framework introduced in previous research which models motivations for exploration and introduce a fitness function for evaluating an environment's potential for exploration. Our study showed that our exploratory agent can clearly distinguish between engaging and unengaging levels. The findings suggest that our agent has the potential to serve as an effective tool for assessing procedurally generated levels, in terms of exploration. This work contributes to the growing field of AI-driven game design by offering new insights into how game environments can be evaluated and optimised for player exploration.
\end{abstract}


\maketitle

\section{Introduction} \label{introduction}

Exploration in video game environments, is an area of study to understand what constitutes engaging and immersive experiences for players. The process of navigating through these virtual environments is often driven by the design of the levels themselves, which can either encourage or hinder exploration. As level designers strive to create more compelling and interactive worlds, understanding the factors that contribute to a good \textbf{exploratory experience} becomes increasingly important.

This study seeks to address the question of what makes certain game levels more conducive to exploration than others. Specifically, it investigates whether levels generated by 2 different procedural content generators (generator A and B) in their ability to facilitate exploration. Generator A is designed to produce levels that are generally engaging, with a balanced navigable area and a variety of interactive elements, while Generator B generates levels that might be considered unengaging, characterised by their lack of objects and object placement throughout a level. Both generators use Wave Function Collapse (WFC) to generate levels. More information about WFC can be found in \ref{wfcbackground}.

To evaluate the effectiveness of these generators, an \textbf{exploratory agent}, modelled after those used in our previous study \cite{khaleque2024experiments}, was employed. The agent's behaviour was analysed based on several key metrics: environment coverage, inspection of unique objects, a custom novelty measure, entropy of the agent’s path and average motivation experienced by the agent on its path. These metrics were chosen to quantify the quality of exploration in a way that balances the novelty and familiarity of the environment, the diversity of pathways, the unpredictability of the agent’s movements, and if the agent is actually finding motivation to explore the environment.

We introduced the concept of an exploratory agent as "a type of agent which traverses a level and explores it in accordance to it's features. It surveys an environment, to observe which features are available in the level, and moves in the direction towards the closest interesting target(s) or direction(s).". 

Exploratory agents have previously been used to evaluate levels in generative systems. For example, in Stahkle et al.'s PathOS framework \cite{stahlke} for assisting designers in level and world design. Cook also investigated evaluating levels with agents that used a vision-based approach \cite{cook_road_2021}, although their project was abandoned. 

Our experiment focuses on the role that exploratory agents can play as part of a PCG framework for differentiating between levels generated through various algorithms, and assessing their suitability for exploration. The key question this study tries to answer is whether exploratory agents are an effective means of filtering procedurally generated levels for exploratory experiences. These agents, by being given possible motivations for exploration, make it possible for developers to get valuable information about the quality and the engagement a level may provide, hence offering a systematic way of optimising PCG processes.

In this paper, we will evaluate the quality of the levels generated with respect to exploratory behaviors of agents by running a number of experiments; these experiments use coverage, entropy, and novelty metrics to see how well these levels support exploration. The long-term aim of the work is to demonstrate that EAs could be both reliable and efficient in filtering and improving the quality of procedurally generated content to ensure that the generated levels are varied, engaging, and support player exploration.

\section{Background} \label{Background}
\subsection{Curiosity Based Exploration} \label{curiosityExploration}
Pathak et al \cite{pathak18largescale} provide an investigation of curiosity-driven learning in artificial agents, with a particular focus on agents operating without any external rewards. The presented work belongs to the area of autonomous agents that learn to control their behaviour in various simulated environments including games and physics simulations, purely driven by intrinsic motivation opertationalised through curiosity. The authors investigated the use of different feature spaces in estimating the prediction error and established that, while random features are enough in some cases, learned features can offer better generalisation. The research pointed to the possible failure of prediction-based rewards, more precisely in the stochastic case, and suggested that additional research on the efficient handling of such environments should be conducted.

Pathak et al's techniques are different from ours because our agent is not meant to be general in the sense that it would explore many different environments using intrinsic motivations. Our agent is given motivations, like the agent in our previous work \cite{khaleque2024experiments}, and it will explore in different ways in different environments.

\subsection{Agents to Assist Game Design} \label{gameDesignAgents}
Stahlke et al \cite{stahlke} introduces PathOS, which predicts player navigation in games that are digital.
Level and world design can be improved by the collected data of how the agents navigate the world. The system was aimed at reducing the burden of playtesting, offering accessibility to devs, being easy to use for designers, and having generalisability. These are similar to the goals we propose with our Exploratory Agents.


A study with 10 participants was carried out using the system, by applying and assessing it; the participants were pre-interviewed, introduced to the system and assigned to create two levels by using the system.

The authors reported that in the post-task interview, impressions of the system as a design tool were quite positive, although the participants mentioned that the behaviour of the agents was very different from how a player would behave.

Nova et al \cite{pathosplus} present PathOS+ as an extension of the basic PathOS framework to complement expert assessments through the simulation of player data by AI. In this way, the problem stated by this strategy aims to deal with the subjectivity of expert assessments by using objective simulated player behaviour data. This would further help improve the reliability of expert assessments among games user research. These are exemplified through the potential of PathOS+, by applying it in a gameplay analysis for navigation and player behaviour.

Furthermore, we \cite{khaleque2024experiments} explored the use of exploratory agents as a method for evaluating hand made game levels based off popular exploratory experiences, particularly in how well these levels support exploration. They pruposed a framework where agents are given motivations in the form of metrics to model motivations for exploration. As we do in this research project. The study showed that different combinations of metrics resulted in distinct exploratory behaviors, which aligns with expectations based on the design of the levels being tested. The researchers demonstrated that such agents could provide valuable feedback for level designers, potentially evaluating and guiding a generative process to create more engaging and exploratory-rich environments. 

\subsection{Wave function Collapse} \label{wfcbackground}
The WFC algorithm, initially inspired by quantum mechanics, has become a prominent tool in the field of procedural content generation (PCG). This algorithm, first used to generate tilemaps, introduced by Maxim Gumin in 2016 \footnote{https://github.com/mxgmn/WaveFunctionCollapse}, operates on the principle of constraint solving and is primarily used to generate patterns or textures that resemble a given input, ensuring that the output adheres to the rules derived from the input data. The WFC algorithm functions by taking an initial grid where each cell can adopt multiple states, much like the quantum superposition, and systematically collapsing these possibilities based on local constraints until a consistent pattern is formed.

WFC, in particular, has gained immense popularity within PCG because of its unique ability to produce coherent and intricate structures from inputs that are relatively simple. This makes it highly relevant in many applications, particularly in game level generation and general scene generation both in 2D and 3D spaces. It excels at the creation of content that remains structurally logical and is thus very suitable when developers need creativity but also coherence in their PCG.

WFC has been applied to a number of Game Development cases in order to automate creation of complex environments with huge variability, hence drastically reducing the burden of manual level design. For instance, it has been used to generate tiling patterns for textures, layouts for dungeon-like environments, and even the infrastructures of virtual worlds. The flexibility provided by WFC can accommodate all these very different content generation scales, small detailed textures versus expansive game worlds, by adjusting the input parameters and the size of the grid used during the generation process.

WFC has been used to generate 3D levels. For example in Bad North \footnote{https://www.badnorth.com/} and by Kleineberg \footnote{https://marian42.itch.io/wfc}. BorisTheBrave also provides a Unity3D package of which allows 3D generation of levels using WFC called Tessera \footnote{https://assetstore.unity.com/packages/tools/level-design/tessera-procedural-tile-based-generator-155425}, which we have decided to use in this research project.

WFC is a powerful and efficient way to implement content creation automation within a videogame. This helps in generating large amounts of content in little time. For this reason, we have chosen to have our generators use WFC.

\section{Agent Metrics}
We \cite{khaleque2024experiments} introduced object and direction based metrics in their framework. We use the same versions of these metrics. Some of them were modified. In this section, we give a brief overview of each metric we used and our modifications. 

\subsection{Direction-Based Metrics}
        
\textbf{Elevation change}: This metric checks if the given direction hits any point in the terrain. If the terrain hit point is higher than the agent's y position (depending on how much higher the hit point is) a maximum value of 1 is given. 0.1 is added until the max of 1 is reached for every unit the terrain hit point is above the agent y. This metric was not modified from our previous version.

\textbf{Openness}: Takes a direction and measures how "open" it is by raycasting checking if there are any objects within a certain distance. Though, boundless space, a raycast hitting nothing is given a score of 0. We previously returned a value of 1 if a raycast hit nothing. However, from a perceptual and gameplay perspective, an environment with no perceivable boundaries might not actually encourage exploration, as there is no clear structure or point of interest to guide the agent’s movement. This can create a sense of aimlessness rather than promoting exploratory behavior.

We have modified this metric to take into account how far an object is, according to the raycast, if the object is as far as the length of view, then 1 is returned otherwise, a fraction representing the percentage of how far the object is in proportion to the view distance (between 0 and 1) is returned.

We have decided to omit the light and shadow, and anticipation direction metrics used by us. This is because measuring varying light intensities was not a goal when generating our levels, and anticipation direction is essentially the same metric as Anticipation object detection, and we decided to go with the object detection version (to which we have renamed Anticipation detection). 

\subsection{Object-Based Metrics}

\textbf{Anticipation Detection}: Is given an object and checks the umbra and penumbra size of the object. It returns a maximum value of 1 and minimum value of 0.  This metric was not modified from our previous version.

\textbf{Large Object Detection}: compares any object with the biggest one it had seen in the course of its run. It returns a value between 0 and 1, indicating how big, in percentage terms, an observed object is relative to the biggest one our agent had ever seen. In case the object is larger than the largest seen so far, then 1 is returned and the largest object seen so far is updated to the most recently seen object. This measure was not modified from our previous version.

\textbf{Group Detection}: Takes an object and checks if there are any other objects in a certain radius (in this case 40 units) of that object. Each object that is close adds a 0.1 to the score, to a maximum of 1. This metric was not modified from our previous version, apart from increasing the radius to check for other objects to account for the size of the assets in our experiment.

We have decided to omit the simple detection metric used in our previous work. We felt this was an overly simplistic metric which did not measure any object properties, as the other metrics do.

\section{Exploratory Agent Framework} \label{explframework}
Our agent framework is very similar to the framework used by in our previous study \cite{khaleque2024experiments}. It uses a system similar to context steering \cite{fray2019context}, in which context maps are formed for each measured direction (36 in total). A context map is a projection of the decision space of the entity onto a 1D array.

Like our previous study's agents there are multiple adjustable parameters:

\textbf{Length of View}: The maximum distance the agent can observe in units

\textbf{Field of View}: The maximum angle of which the agent can observe, independent of the camera attached

\textbf{Decision Time}: The time step to recalculate the interest map and direction to move in for the agent.

In this framework, interest maps are formed from a list of objects which are in view of the agent. A camera is used to detect which objects are in view and only samples directions within view of its camera. There are 36 directions sampled in total. The highest scoring direction is chosen to be moved in. 

There are three main stages to the pipeline, explained below.

\textbf{Stage 1: Selecting a subset of objects} A camera is used to survey the surrounding area of the agents. Every object within the camera frustrum is added to a list representing the objects of interest. The output from this stage is a list of objects of interest.

\textbf{Stage 2: Making Interest Judgments} The list from stage 1 is taken and an interest map, consisting of a score associated with a direction is formed. For each direction, a direction based metric is applied to calculate the directions interest score. Also, object based metrics are applied, each object has it's direction taken and rounded to the closest direction in the direction interest map (and added to the direction interest map) before the direction score is updated. 

\textbf{Stage 3: Making a Navigation Decision} The direction map of Stage 2 is used to make a navigation decision. The direction of highest interest is chosen. If there is an object which is associated with the highest scoring direction, that is chosen to be navigated towards using A* pathfinding \footnote{https://www.arongranberg.com/astar/}. Objects are only associated with directions when an object based metric is being used. A direction multiplied by 50 steps is chosen to be moved in. In our previous work \cite{khaleque2024experiments} use the unity navmesh system and move 10 steps in the direction of the highest scoring direction we chose to make these changes because moving in larger steps, the agent commits more strongly to the direction identified as the most promising based on its internal metrics. This commitment ensures that the agent fully explores the opportunities presented by high-scoring directions, maximizing the benefits of moving towards areas that offer the greatest potential for novelty, object interaction, or other desirable outcomes. If there are multiple directions that are scored as the highest, a random one is chosen.


\section{Generator Details} \label{GenDetail}
As mentioned before, our generators use WFC to generate level. We made these generators in the Unity game engine \footnote{https://unity.com/}, using Tessera \footnote{https://assetstore.unity.com/packages/tools/level-design/tessera-procedural-tile-based-generator-155425}. Details of each generator are given in the following subsection. 

Each generator has 35 tiles, each of these 35 tiles have a chance (a float between 0 and 1) of being spawned. Each generator can generate a level of 350x350 units, each tile is 50 x 50, so 49 tiles can be generated to form a level. Each of these tiles can have 4 x rotation(0, 90, 180 and 270). The theoretical number of possible levels is (35x4)\textsuperscript{49} levels for both of these generators. 

However, because each tile has a probability associated with it, the effective expressive range is influenced by these probabilities. This means that the actual number of levels that can be meaningfully generated might be less.

\textbf{Generator A} uses slightly higher probabilities for decorated tiles and tiles with elevation/slopes, increasing the chance for a large variety of objects in our engaging levels.

\textbf{Generator B} uses slightly lower probabilities for decorated tiles and tiles with elevation/slopes, decreasing the chance for a large variety of objects in our engaging levels, leading to emptier levels/levels decorated with the same types of objects and/or less elevated positions. Generator A and B both use the same tileset.

Exploratory agents are capable of assessing how the spatial arrangement and layout of the levels influence exploration. This includes the distribution of objects which might attract players to explore, or the complexity of paths. Differentiating levels through other means, e.g. simply object counting, does not account for these spatial relationships, which are crucial for understanding the navigability and engagement level of a game environment. So, even though generator B, on average, produces levels with fewer objects, simple methods to determine whether a level is engaging or not would not necessarily be effective.

In addition, simpler methods of evaluating levels provide a raw measure of quantity but do not account for the spatial distribution, contextual significance, or interactive potential of these objects. Such evaluation criteria are limited in their ability to reflect the complexity of the player experience, where placement, accessibility, and interaction opportunities within the level are crucial determinants of exploration quality. Exploratory agents, on the other hand, offer a dynamic evaluation of levels by simulating motivations for exploration, allowing them to assess not just what is present in the environment, but how it might be experienced during exploration.


\section{Evaluating Generated Levels} \label{evalLevels}

In order to demonstrate how well an environment might support exploration, we decided to use the evaluation criteria we used previously (Coverage, object inspection and novelty) as well as expand on it by adding our own. Our additions to the evaluation criteria include; Entropy, modifications to the novelty measure and measuring agent motivation over time. Also, like our previous work, we measured agent trajectory for each metric and spawnpoint \footnote{https://github.com/BKhaleque/Evaluating-Environments-using-Exploratory-Agents}. Using the mentioned evaluation criteria, we created a fitness function to give a score (between 0 and 1) for each level.

\subsection{Coverage}
Coverage serves as a measure of inspective exploration, as described in our previous work \cite{khaleque2024experiments}, we used the same technique to derive coverage as they did. Simply by counting how many of the 50x50 regions the agent had visited with their respective metrics.

We expect less coverage on average in our unengaging levels than in our engaging levels. Our unengaging levels lack a lot of the stimuli that drive an agent to explore the environment fully. Engaging levels often have more of these interactive elements, such as a wider variety of objects and more large objects, which motivate the agent to traverse the entire space. In contrast, unengaging levels are more repetitive, offering little to no reward for thorough exploration. As a result, the agent may not cover a lot of the area of the unengaging levels, leading to reduced coverage.

\subsection{Inspection}
Another measure of inspective exploration. This was a measure (in terms of percentage) of how many objects were seen and visited by the agent. We also used the same technique as Khaleqe et al.  We measured the percentage of objects the agent came within 10 units of. A higher inspection score suggests that the agent had a "want" to learn about the objects in the environment whereas a lower one suggests the opposite.

We expect our agent to have a lower inspection score in our engaging levels than in our unengaging levels. Our unengaging levels lack a lot of the diversity and complexity found in engaging levels, which can lead to a higher concentration of the few available objects in the agent’s field of view. Since there are fewer distinctive or appealing areas to explore, the agent might spend more time interacting with the objects it encounters, leading to a higher object inspection score. In contrast, in engaging levels, the agent might be more drawn to explore the environment as a whole rather than focusing on individual objects.

\subsection{Entropy}
This is a measure of Shannon entropy \cite{shannonentropy}. Shannon entropy, a foundational concept in information theory introduced by Claude Shannon in 1948, quantifies the uncertainty or unpredictability of a random variable. For a discrete random variable \( X \) with possible outcomes \( \{x_1, x_2, \dots, x_n\} \) occurring with probabilities \( \{p_1, p_2, \dots, p_n\} \), the Shannon entropy \( H(X) \) is defined as:

\begin{equation}
H(X) = -\sum_{i=1}^{n} p(x_i) \log_2 p(x_i)
\end{equation}

Here:
\begin{itemize}
    \item \( p(x_i) \) is the probability of outcome \( x_i \),
    \item \( \log_2 p(x_i) \) is the logarithm of the probability in base 2, giving the entropy in bits.
\end{itemize}

Shannon entropy achieves its maximum value when all outcomes are equally likely, which corresponds to maximum uncertainty. Conversely, if one outcome is certain (i.e., \( p(x_i) = 1 \) for some \( i \) and 0 for others), the entropy becomes zero, indicating no uncertainty.

In our study, Shannon entropy is used to measure the diversity and unpredictability of the agent’s exploration path within the generated levels. The core reason for using Shannon entropy in this context is its ability to quantify the randomness of the agent's movements, which reflects the variety of choices the agent makes during exploration.

By calculating the entropy over the grid locations visited by the agent, we can determine whether the agent's path was too predictable (low entropy) or too random (high entropy). An ideal exploration path strikes a balance, showing neither excessive randomness nor predictability, which is crucial for maintaining engagement in exploratory experiences.

We expect our engaging levels to have less entropy, on average, than our unengaging levels. This is because our engaging levels have more meaningful object placements with a wider spread and a variety of objects that we expect will attract the agent's attention with any given metric causing more focused exploration in the engaging levels reducing the randomness of the agent's path.

\subsection{Novelty}
We used novelty measured over each time step to help evaluate the generated environments. Novelty is a measure of the stimuli experienced by the agent in it's path. We use a very similar measure to our previous work, though instead of measuring the novelty at each 50x50 region, we measure the novelty experienced by the agent at each time step (1 second).

The novelty score can be explained as follows:

\begin{itemize}
    \item \( N_t \) represent the novelty score at time \( t \) for a given type of object.
    \item \( S_t \) represent the total novelty score  \( t \).
    \item \( \Delta t \) represent the time interval, where \( \Delta t = 0.1 \) seconds.
    \item \( r \) represent the rate of novelty score recovery, where \( r = 0.03 \) per second.
    \item \( M \) represent the maximum novelty score an object type can recover to, where \( M = 0.1 \).
    \item \( P \) represent the penalty applied to the novelty score when an object type is seen, where \( P = 0.1 \).
    \item \( v_t \) represent the visibility flag at time \( t \), where \( v_t = 1 \) if the object type is seen and \( v_t = 0 \) otherwise.
\end{itemize}

\section*{Initial Condition}
When a type of object is encountered for the first time:
\[
N_0 = M = 0.1
\]
and it is marked as "seen".

\section*{Novelty Score Update}
The novelty score at time \( t \) is updated as follows:
\begin{itemize}
    \item If the object type is \textbf{not seen} (\( v_t = 0 \)):
    \[
    N_{t + \Delta t} = \min(N_t + r \cdot \Delta t, M)
    \]
    \item If the object type is \textbf{seen} (\( v_t = 1 \)) and it is "new":
    \[
    N_{t + \Delta t} = N_t - P
    \]
    \item If the object type is \textbf{seen} (\( v_t = 1 \)) and it is not "new":
    \[
    N_{t + \Delta t} = N_t + r \cdot \Delta t
    \]
\end{itemize}

When the object type is seen at time \( t \), the total novelty score is updated:
\[
S_t = S_t + N_t
\]

\begin{itemize}
    \item The novelty score \( N_t \) for an object type starts at \( 0.1 \) when the object is first seen.
    \item Each time step \( \Delta t = 0.1 \) seconds, the score either recovers at \( 0.01 \) per second (if the object type is not seen), or is penalized by \( 0.1 \) (if it is seen for the first time).
    \item The tile score \( S_t \) accumulates the novelty score \( N_t \) of each object type seen at that time.
\end{itemize}

We expect novelty to be lower in our unengaging levels due to a lack of diverse elements and a more simplisitic level design. When the environment lacks diversity, the agent quickly becomes familiar with the surroundings, leading to a decrease in perceived novelty. 
\subsection{Motivation}
Motivation is a measure of the highest scoring direction, according to the agent's attached metrics, at a given time step (1 second). The motivation metric serves as an indirect measure of how well the environment supports exploration. If the agent consistently finds high motivation, it suggests that those areas are well designed to encourage exploration. Conversely, if the agent's motivation decreases, it may indicate that the environment lacks sufficient stimuli or that the design is too predictable, leading to disengagement.

\subsection{Fitness Function for Evaluating Levels}

To evaluate the exploratory potential of generated levels, we designed a fitness function that integrates all the evaluation criteria mentioned above, each reflecting different aspects of the agent's exploration. The fitness function is structured as follows:

\( F \) is the overall fitness score of a level, calculated as:

\begin{equation}
F = \sum w \cdot f_m
\end{equation}

where:
\begin{itemize}
    \item \( w \) is the weight given to a respective metric
    \item \( f_m \) is the fitness score for the metric.
\end{itemize}

The fitness score for each metric is determined by the following criteria:

\begin{enumerate}
    \item Coverage : The fitness score \( f_m \) is set to 0 if the average coverage over all spawns falls outside the range of 20\% to 80\%. If the coverage is within this range, \( f_c \) is multiplied by the product of average motivation (\( M_{\text{avg}} \)) and average novelty (\( N_{\text{avg}} \)). Ensuring that coverage is between 20\% and 80\% prevents the extremes where too little exploration indicates a sparse or uninteresting level, and too much coverage might suggest that the level is too small or lacks sufficient depth to sustain interest.

    \item Entropy: The fitness score \( f_m \) is set to 0 if the average entropy over all spawns exceeds 0.9. Limiting entropy to 0.9 ensures that the agent's exploration is too chaotic. High entropy might indicate that the level design is overly complex or lacks clear direction, which could detract from the player experience.

    \item Object Inspection: The fitness score \( f_i \) is set to 0 if the average object inspection over all spawns exceeds 80\%. If the inspection rate is 80\% or lower and greater than 10\%, \( f_i \) is multiplied by \( M_{\text{avg}} \cdot N_{\text{avg}} \). Capping object inspection at 80\% prevents the agent from being overly focused on objects, which might indicate that the level is too cluttered or lacks broader exploratory opportunities and having at least 10\% of objects inspected makes sure there are at least some objects that are worth being investigated. This balance ensures that the level is engaging both at the micro and macro levels.

    \item Motivation and Novelty: Both average motivation over all spawns and average novelty over all spawns are directly factored into the multiplication to emphasize the importance of these metrics in evaluating the exploratory potential of the level. Multiplying each metric's score by the average motivation and novelty ensures that the fitness function favors levels that actively encourage exploration and offer new experiences. High motivation suggests that the level engages the agent effectively, while high novelty indicates that the level provides new stimuli and avoids repetition.

\end{enumerate}

The final fitness score \( F \) therefore balances the contributions of individual exploration metrics with the comprehensive assessment provided by the agent loaded with all metrics. The thresholds and multipliers ensure that levels only receive a high fitness score if they meet essential criteria for meaningful exploration, such as balanced coverage, manageable entropy and appropriate object inspection.

\section{Experiment Setup}\label{topdown-engaging}
To evaluate procedurally generated environments using \textbf{exploratory agents}, we conducted a study with our agent exploring 5 levels generated from generator A, considered engaging and 5 levels from generator B, considered unengaging. We looked at trajectories of the agent using the above metrics as well as an agent with a combination of all the metrics previously mentioned to gather data on how multiple metrics explore both engaging and unengaging levels. All levels were the same size (350x350) units.

We ran our agent with each metric for 3 minutes at 3 different spawn points for each of the levels. We also had a random control agent for which we measured coverage, novelty, entropy and object inspection for. We did not measure motivation for this agent as it didn't have the capability to have a metric loaded onto it. 

The limited length and field of view does mean the spawn point will likely greatly affect the agent paths; to obtain a broader sample, we tested 3 different spawn points on 3 levels.

In our experiment, we tested all singular metrics before loading an agent with every combined metric. This approach was taken to ensure a comprehensive understanding of how each metric influences the agent's behaviour and the overall exploration process. 

By first testing each metric individually, we observed the specific influence each one had on the agent and how it influenced the agents decision-making and exploration patterns.

We decided to load an agent with all metrics simultaneously to create a comprehensive and balanced exploratory agent that we thought might effectively navigate complex, procedurally generated environments and provide the most useful data in identifying engaging and unengaging levels.

For the experiment the following agent parameters were set to the values described below:

\textbf{Length of View}: 115. Previously we set this to 80. Through preliminary testing of our agents we found this was too low of a value to observe many of the objects in these types of generated levels. 115 was a much better value in this regard

\textbf{Field of View}: 90. This was observed to be an appropriate value. Like our previous study, we thought a 90 degree FOV provided a good angle to perceive objects. It is standard in many first person games and does not allow an excessive view of the environment

For our fitness function, every metric was given a weight of 0.1, except for our agent which had all the metrics loaded in, this combination of metrics was given a weight of 0.5 This was because the combined metrics represent a holistic evaluation of the level's exploratory potential, integrating the insights provided by all the individual metrics. The agent that incorporates all these metrics is likely to offer a more accurate and comprehensive assessment of the level's quality for exploration, capturing the interplay between different motivations for exploration that might be missed when metrics are considered in isolation.

By giving more weight to the combined metrics, the fitness function emphasises the importance of a balanced exploration experience, where all factors are considered in tandem rather than in isolation. This approach ensures that levels which perform well across all dimensions of exploration are more highly rated, reflecting the belief that such levels are more likely to provide a rich, engaging experience for exploration. 

\begin{figure}[p] \label{topdown-engaging}
\begin{subfigure}{.45\linewidth}
  \includegraphics[width=\linewidth, height = 2.8cm]{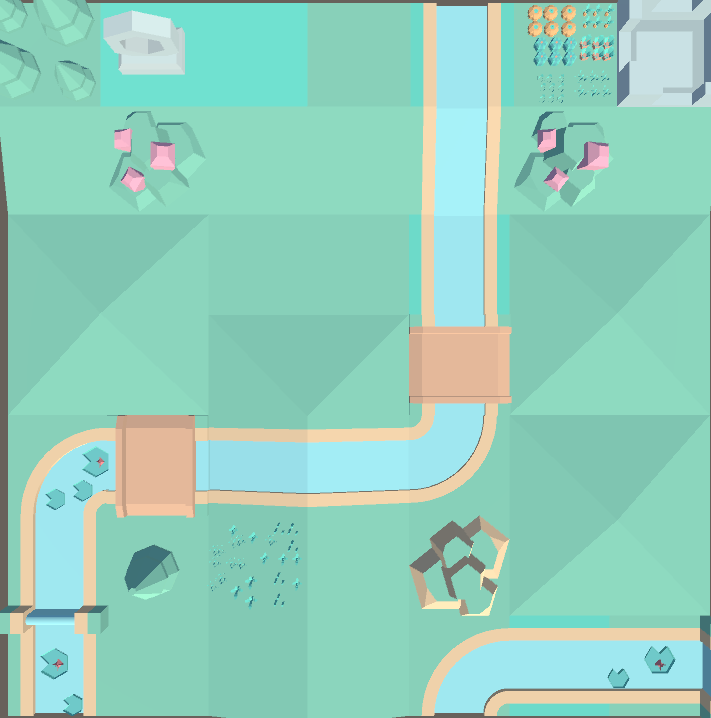}
  \caption{Level 1}
\end{subfigure}\hfill
\begin{subfigure}{.45\linewidth}
  \includegraphics[width=\linewidth, height = 2.8cm]{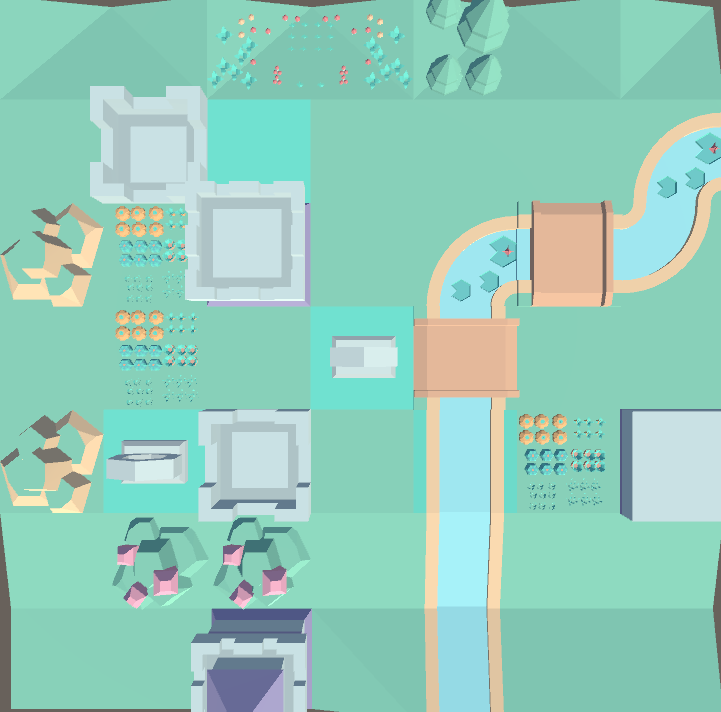}
  \caption{Level 2}
\end{subfigure}\hfill
\medskip 
\begin{subfigure}{.45\linewidth}
  \includegraphics[width=\linewidth, height = 2.8cm]{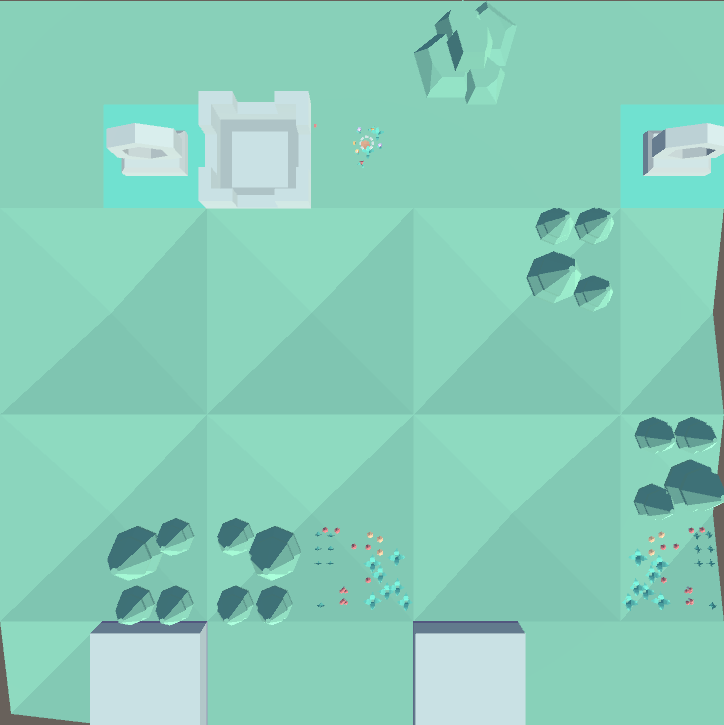}
  \caption{Level 3}
\end{subfigure}\hfill 
\begin{subfigure}{.45\linewidth}
  \includegraphics[width=\linewidth,height = 2.8cm]{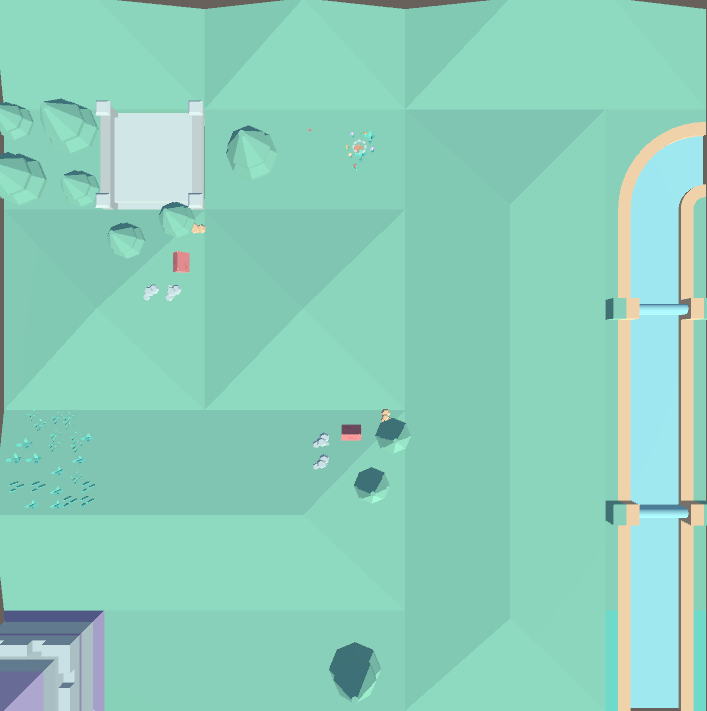}
  \caption{Level 4}
\end{subfigure}\hfill 
\begin{subfigure}{.45\linewidth}
  \includegraphics[width=\linewidth, height = 2.8cm]{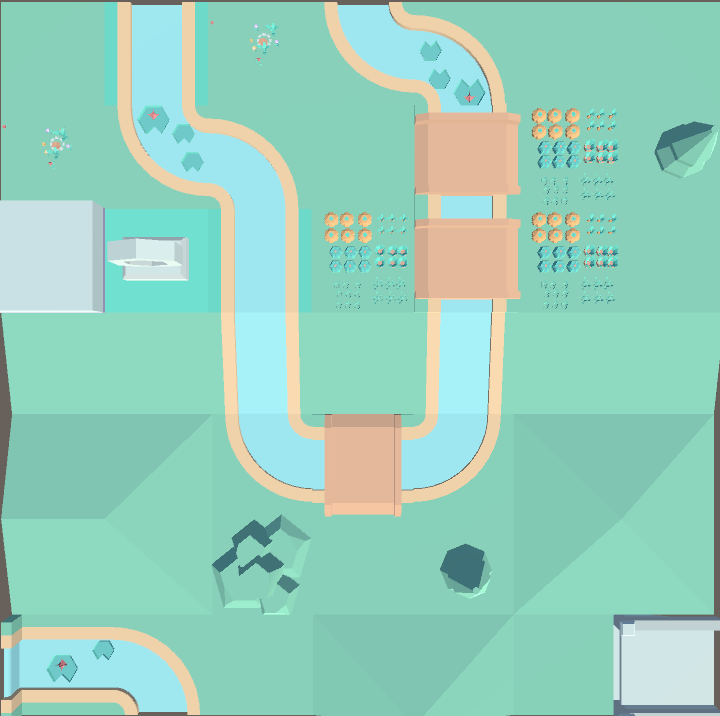}
  \caption{Level 5}
\end{subfigure}\hfill 

\caption{Top down views of all the engaging levels}

\end{figure}

\begin{figure}[p] \label{topdown-unengaging}
\begin{subfigure}{.45\linewidth}
  \includegraphics[width=\linewidth, height = 2.8cm]{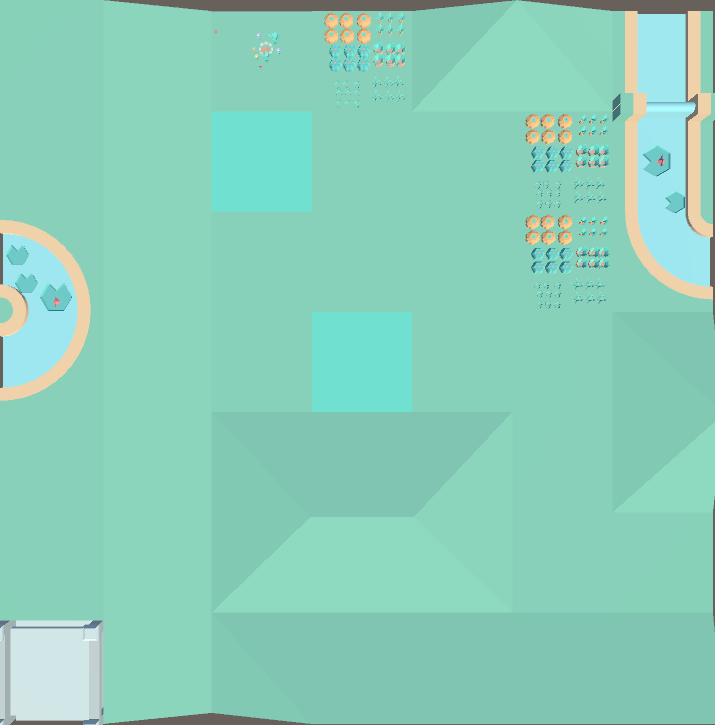}
  \caption{Level 1}
\end{subfigure}\hfill
\begin{subfigure}{.45\linewidth}
  \includegraphics[width=\linewidth, height = 2.8cm]{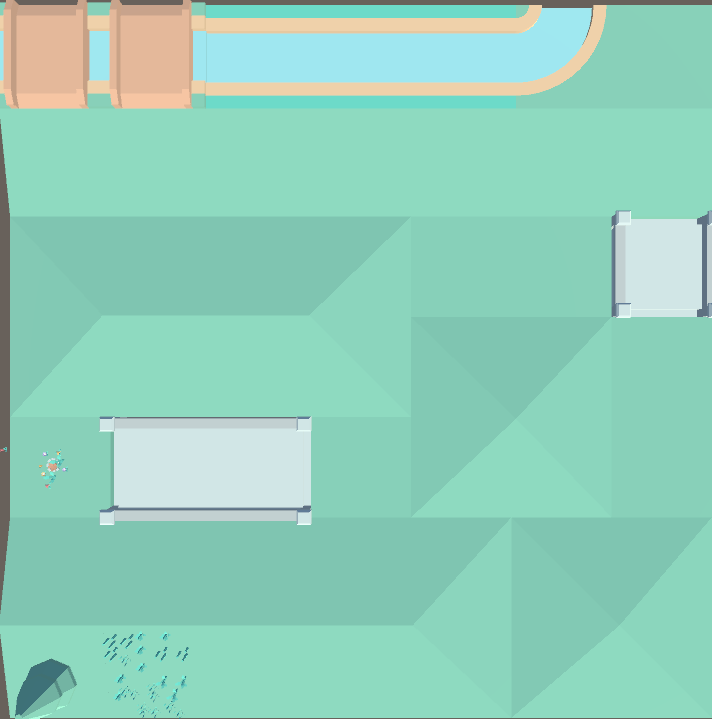}
  \caption{Level 2}
\end{subfigure}\hfill
\medskip 
\begin{subfigure}{.45\linewidth}
  \includegraphics[width=\linewidth, height = 2.8cm]{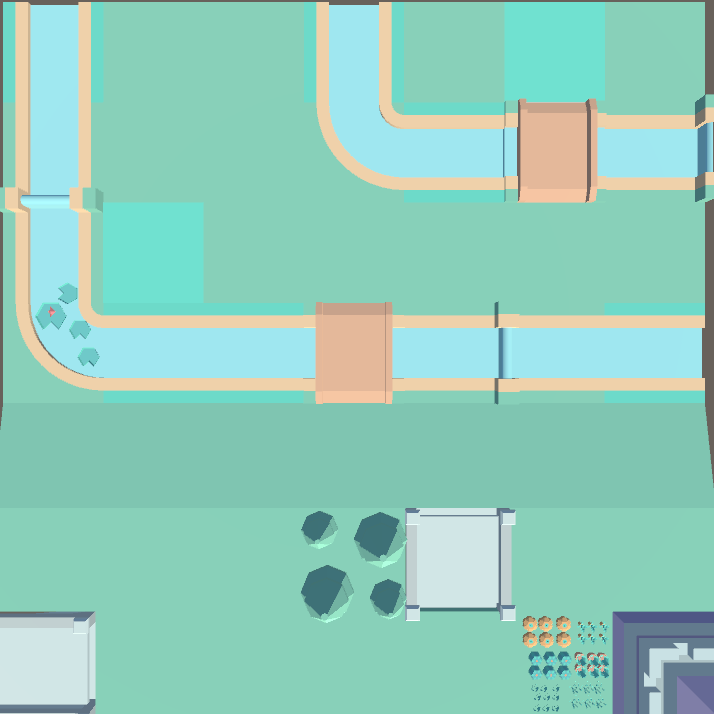}
  \caption{Level 3}
\end{subfigure}\hfill 
\begin{subfigure}{.45\linewidth}
  \includegraphics[width=\linewidth,height = 2.8cm]{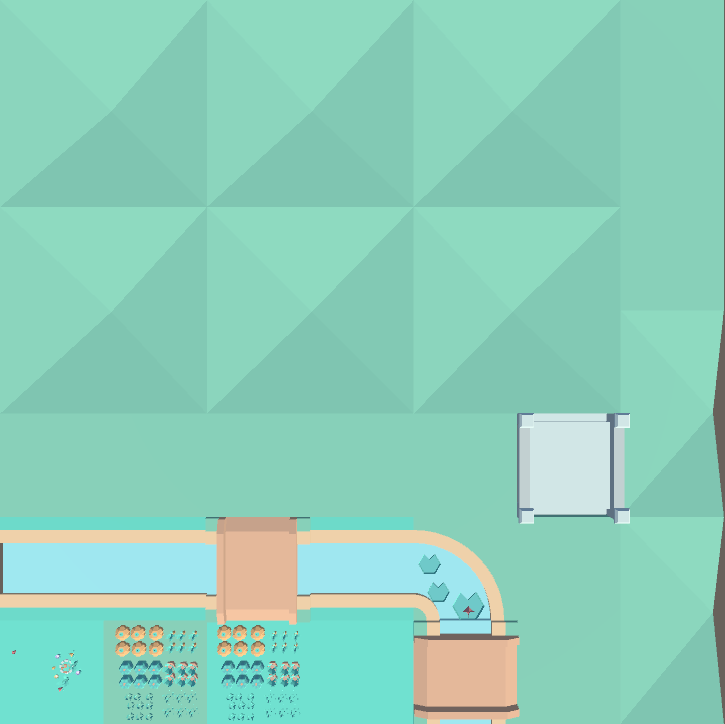}
  \caption{Level 4}
\end{subfigure}\hfill 
\begin{subfigure}{.45\linewidth}
  \includegraphics[width=\linewidth, height = 2.8cm]{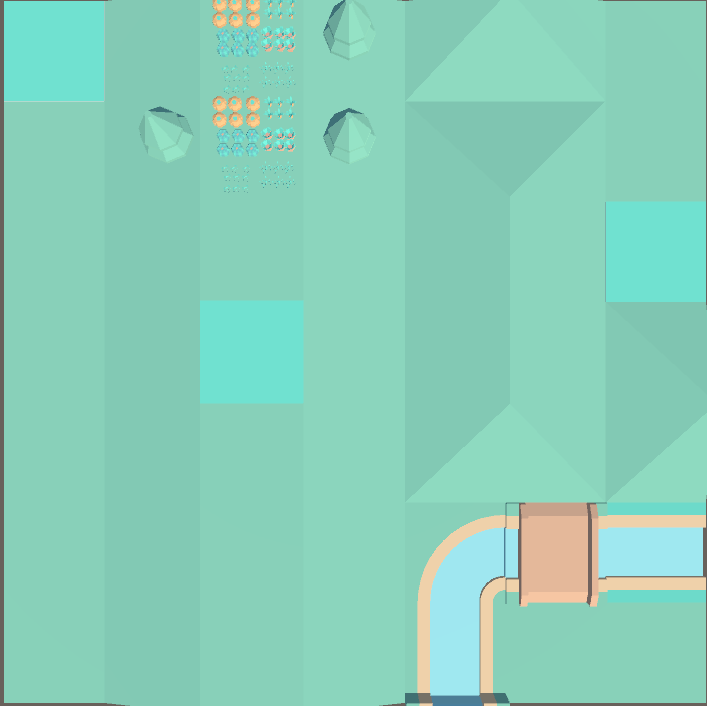}
  \caption{Level 5}
\end{subfigure}\hfill 

\caption{Top down views of all the unengaging levels}

\end{figure}
\section{Results} \label{results}
\begin{figure}[p] \label{motivation-engaging}
\begin{subfigure}{.45\linewidth}
  \includegraphics[width=\linewidth, height = 2.8cm]{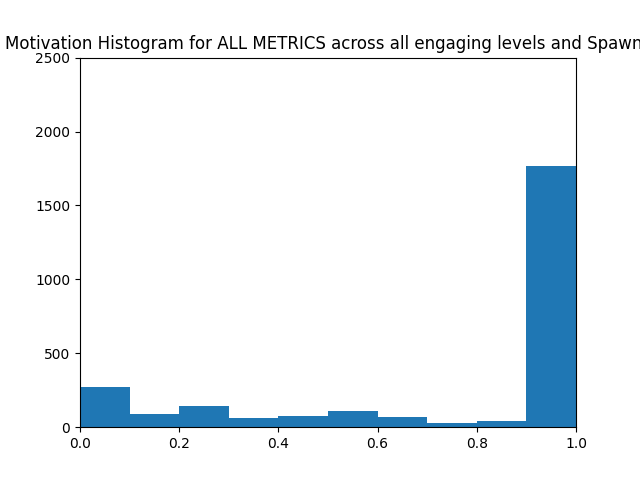}
  \caption{All metrics}
\end{subfigure}\hfill
\begin{subfigure}{.45\linewidth}
  \includegraphics[width=\linewidth, height = 2.8cm]{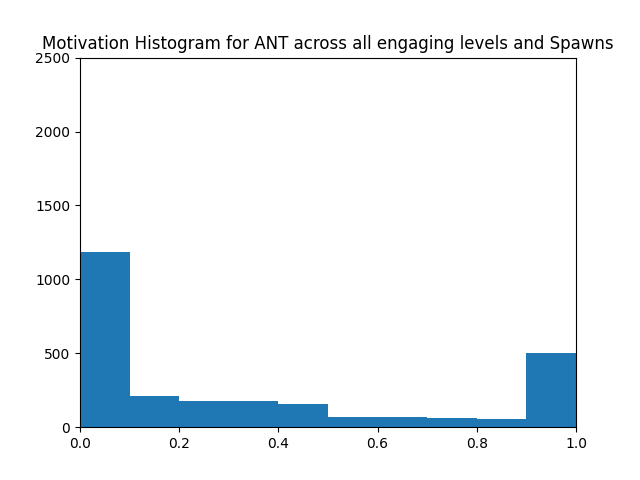}
  \caption{Anticipation Detection}
\end{subfigure}\hfill
\medskip 
\begin{subfigure}{.45\linewidth}
  \includegraphics[width=\linewidth, height = 2.8cm]{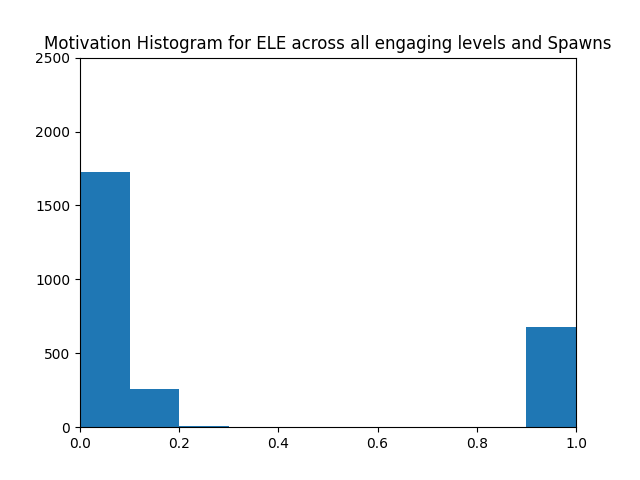}
  \caption{Elevation Change}
\end{subfigure}\hfill 
\begin{subfigure}{.45\linewidth}
  \includegraphics[width=\linewidth,height = 2.8cm]{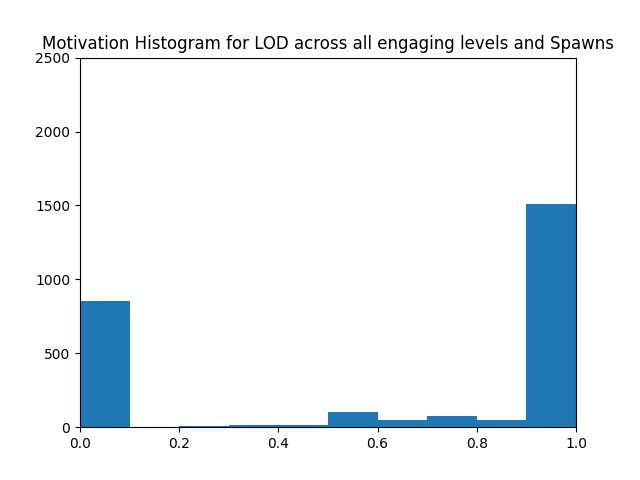}
  \caption{Large object detection}
\end{subfigure}\hfill 
\begin{subfigure}{.45\linewidth}
  \includegraphics[width=\linewidth, height = 2.8cm]{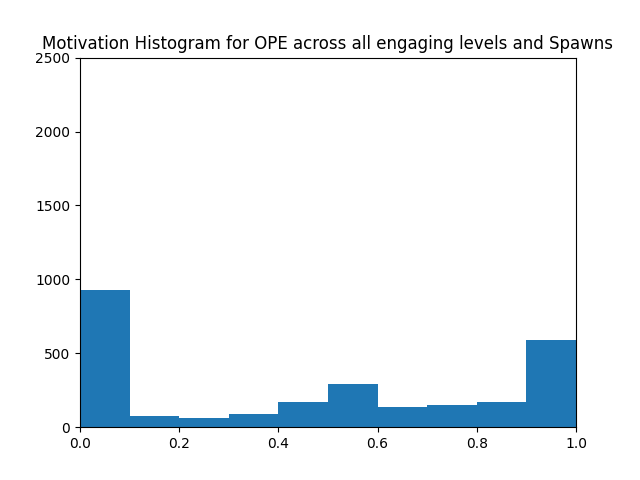}
  \caption{Openness}
\end{subfigure}\hfill 
\begin{subfigure}{.45\linewidth}
  \includegraphics[width=\linewidth, height = 2.8cm]{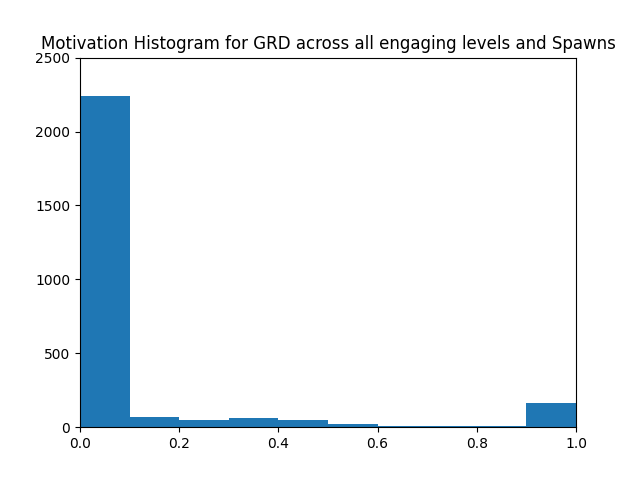}
  \caption{Group Detection}
\end{subfigure}\hfill 
\caption{Motivation histograms for all 3 spawns for all the engaging levels}
\label{topdown}
\end{figure}

\begin{figure}[p] \label{motivation-unengaging}
\begin{subfigure}{.45\linewidth}
  \includegraphics[width=\linewidth, height = 2.8cm]{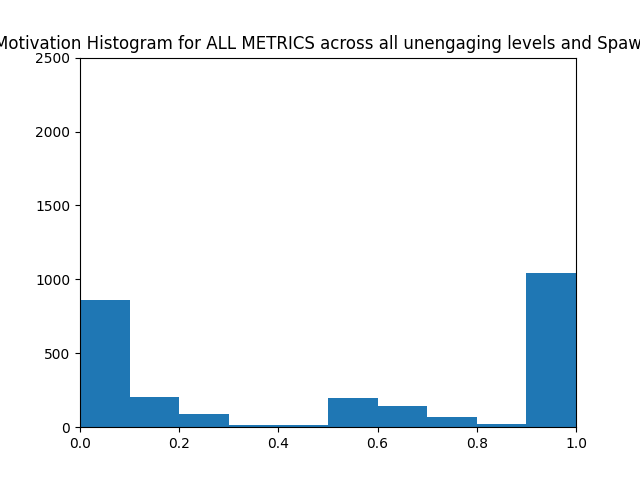}
  \caption{All metrics}
\end{subfigure}\hfill
\begin{subfigure}{.45\linewidth}
  \includegraphics[width=\linewidth, height = 2.8cm]{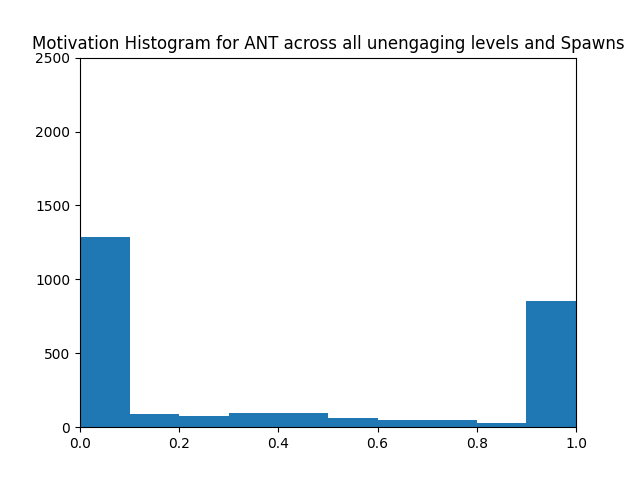}
  \caption{Anticipation Detection}
\end{subfigure}\hfill
\medskip 
\begin{subfigure}{.45\linewidth}
  \includegraphics[width=\linewidth, height = 2.8cm]{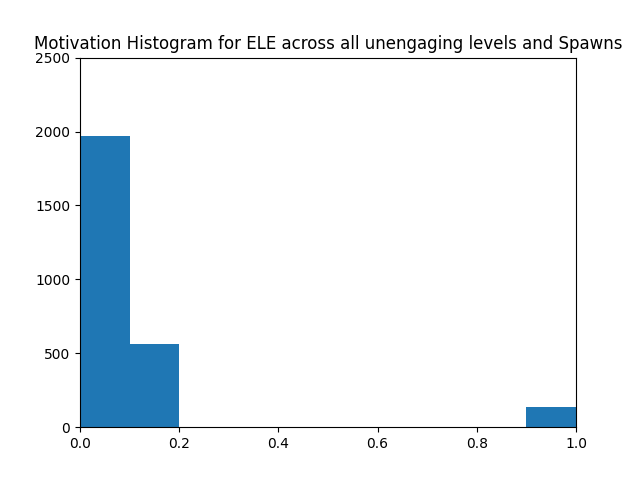}
  \caption{Elevation Change}
\end{subfigure}\hfill 
\begin{subfigure}{.45\linewidth}
  \includegraphics[width=\linewidth,height = 2.8cm]{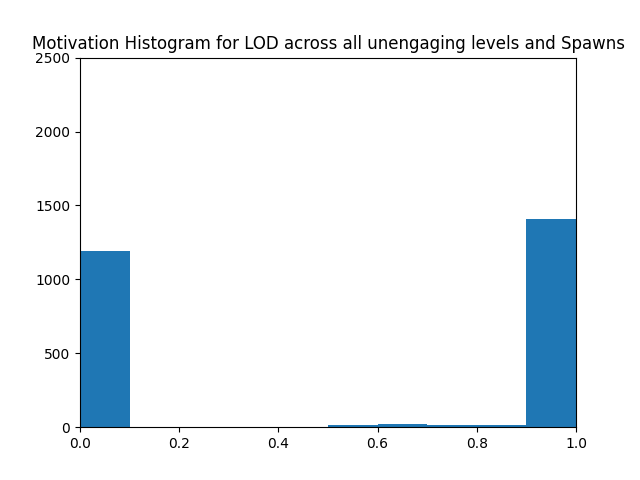}
  \caption{Large object detection}
\end{subfigure}\hfill 
\begin{subfigure}{.45\linewidth}
  \includegraphics[width=\linewidth, height = 2.8cm]{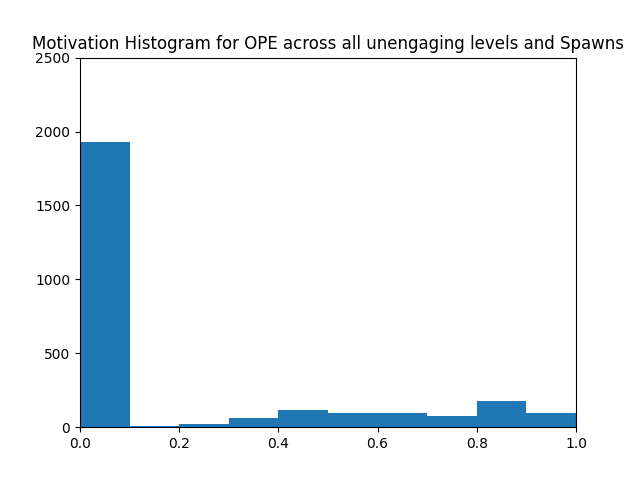}
  \caption{Openness}
\end{subfigure}\hfill 
\begin{subfigure}{.45\linewidth}
  \includegraphics[width=\linewidth, height = 2.8cm]{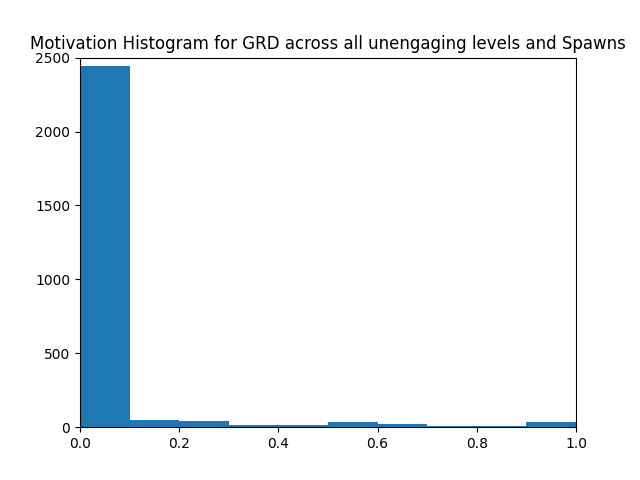}
  \caption{Group Detection}
\end{subfigure}\hfill 

\caption{Motivation histograms for all 3 spawns for the unengaging levels}
\label{topdown}
\end{figure}

The motivation histogram for the unengaging levels vs the engaging levels shows much higher motivation frequencies, on average, for every metric tested, on the engaging levels rather than the unengaging levels. This shows that the paths followed by each metric were considered interesting, at least, much more interesting than our unengaging levels. This also suggests that there were more objects/phenomena that were interesting compared to the unengaging levels.

There were larger friquencies of low or 0 motivation (along with frequencies to high motivation due to how the metric works), particularly for large object detection and openness, in the engaging levels compared to the unengaging levels; this is due to the unengaging levels having a smaller spread of objects throughout their levels, so the agent was spending more time in particular areas of the map where most of the objects were placed. This is also evidenced by the trajectory plots \footnote{https://github.com/BKhaleque/Evaluating-Environments-using-Exploratory-Agents}.

\begin{figure}[t] \label{novelty-engaging}
\begin{subfigure}{.45\linewidth}
  \includegraphics[width=\linewidth, height = 2.8cm]{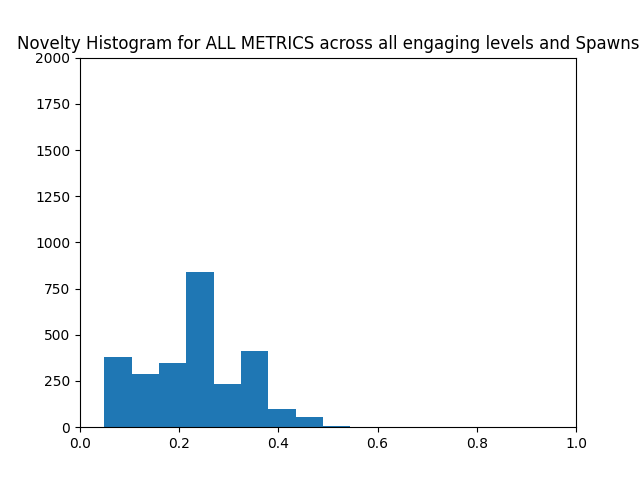}
  \caption{All metrics}
\end{subfigure}\hfill
\begin{subfigure}{.45\linewidth}
  \includegraphics[width=\linewidth, height = 2.8cm]{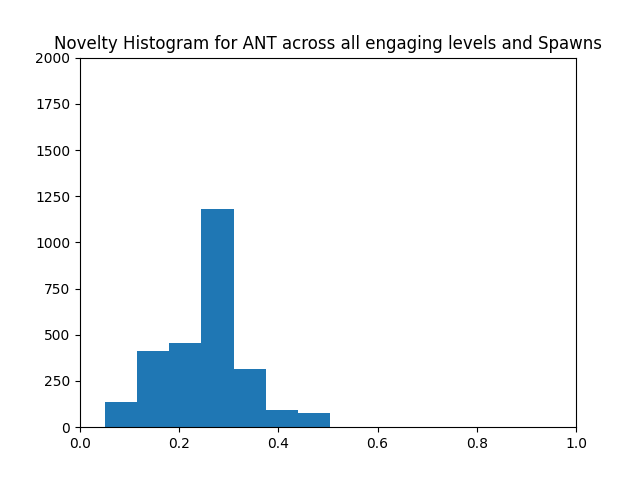}
  \caption{Anticipation Detection}
\end{subfigure}\hfill
\medskip 
\begin{subfigure}{.45\linewidth}
  \includegraphics[width=\linewidth, height = 2.8cm]{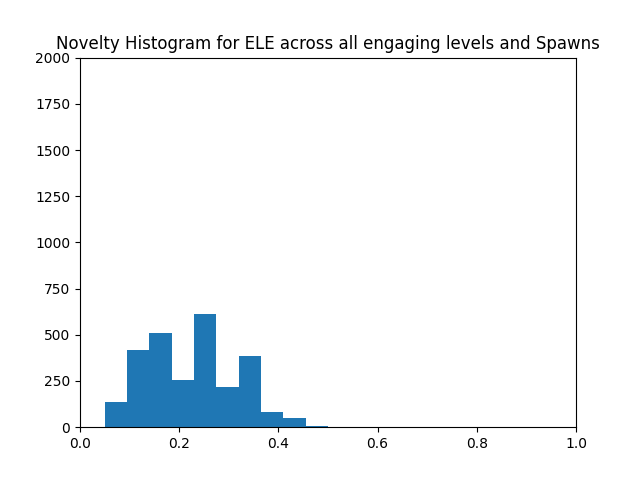}
  \caption{Elevation Change}
\end{subfigure}\hfill 
\begin{subfigure}{.45\linewidth}
  \includegraphics[width=\linewidth,height = 2.8cm]{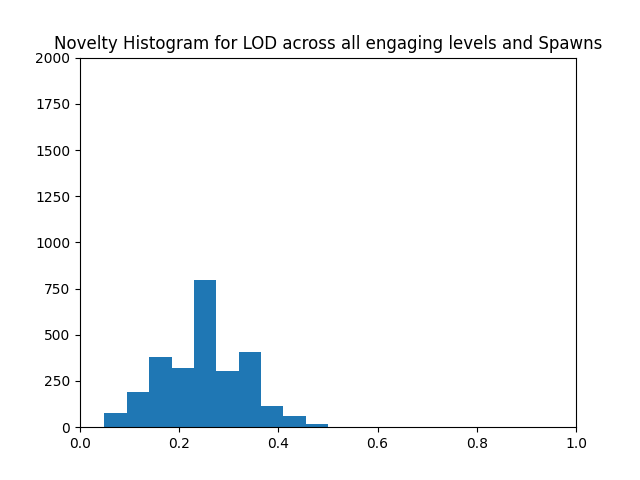}
  \caption{Large object detection}
\end{subfigure}\hfill 
\begin{subfigure}{.45\linewidth}
  \includegraphics[width=\linewidth, height = 2.8cm]{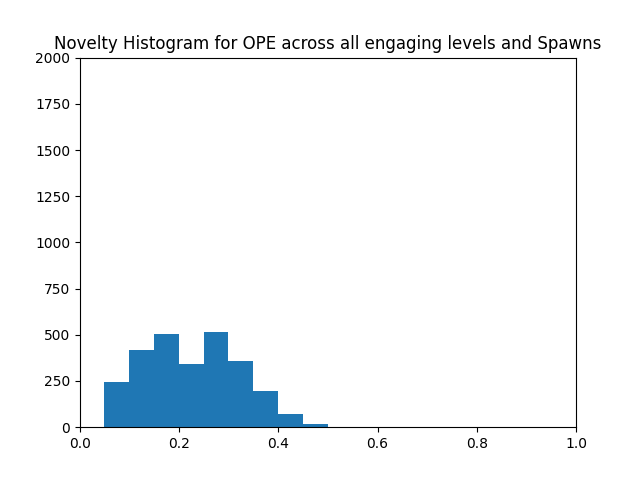}
  \caption{Openness}
\end{subfigure}\hfill 
\begin{subfigure}{.45\linewidth}
  \includegraphics[width=\linewidth, height = 2.8cm]{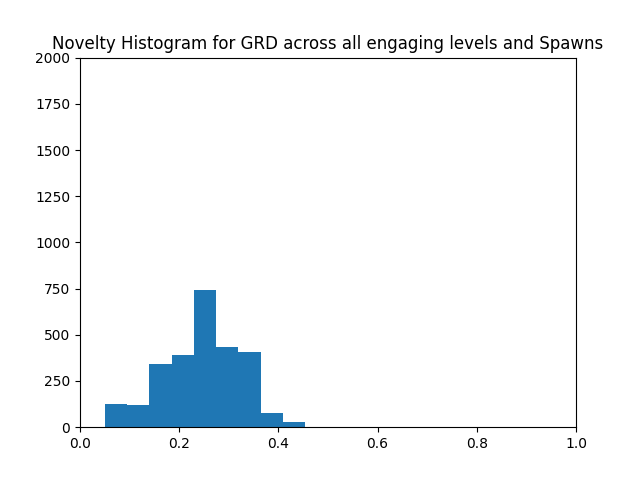}
  \caption{Group Detection}
\end{subfigure}\hfill 
\begin{subfigure}{.45\linewidth}
  \includegraphics[width=\linewidth, height = 2.8cm]{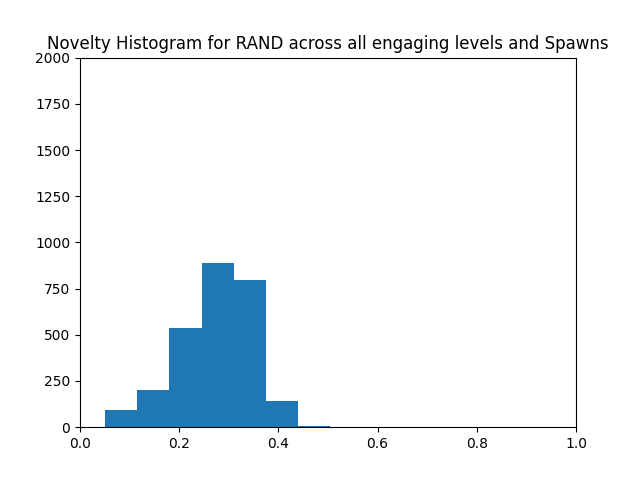}
  \caption{Random Agent}
\end{subfigure}\hfill 

\caption{Novelty Histrograms for all 3 spawns for the engaging levels}
\label{topdown}

\end{figure}

\begin{figure}[t] \label{novelty-unengaging}
\begin{subfigure}{.45\linewidth}
  \includegraphics[width=\linewidth, height = 2.8cm]{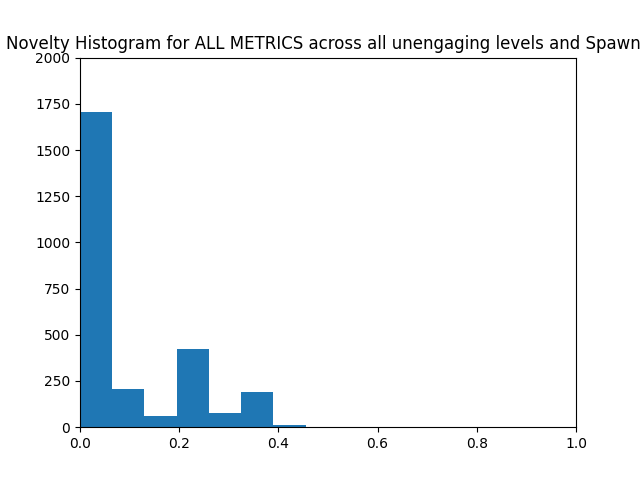}
  \caption{All metrics}
\end{subfigure}\hfill
\begin{subfigure}{.45\linewidth}
  \includegraphics[width=\linewidth, height = 2.8cm]{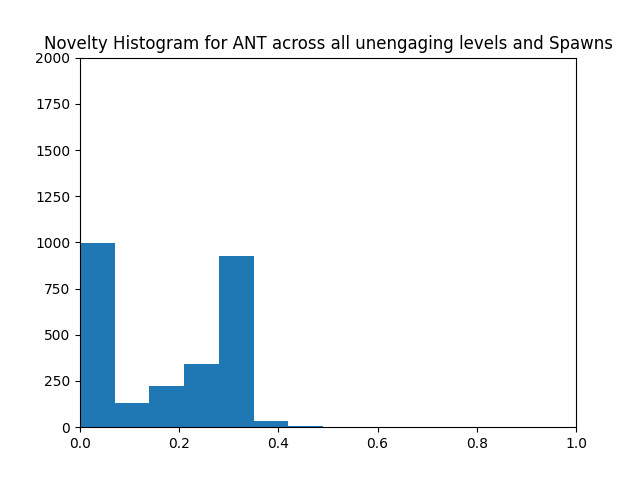}
  \caption{Anticipation Detection}
\end{subfigure}\hfill
\medskip 
\begin{subfigure}{.45\linewidth}
  \includegraphics[width=\linewidth, height = 2.8cm]{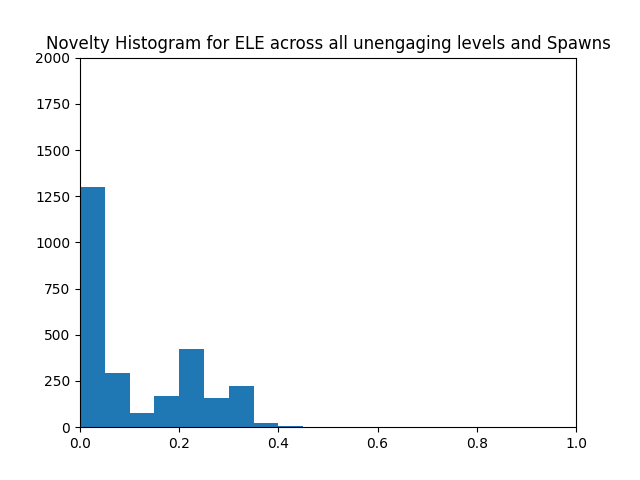}
  \caption{Elevation Change}
\end{subfigure}\hfill 
\begin{subfigure}{.45\linewidth}
  \includegraphics[width=\linewidth,height = 2.8cm]{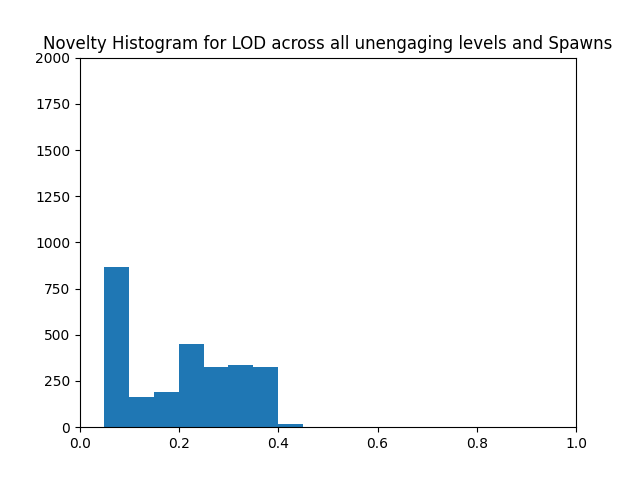}
  \caption{Large object detection}
\end{subfigure}\hfill 
\begin{subfigure}{.45\linewidth}
  \includegraphics[width=\linewidth, height = 2.8cm]{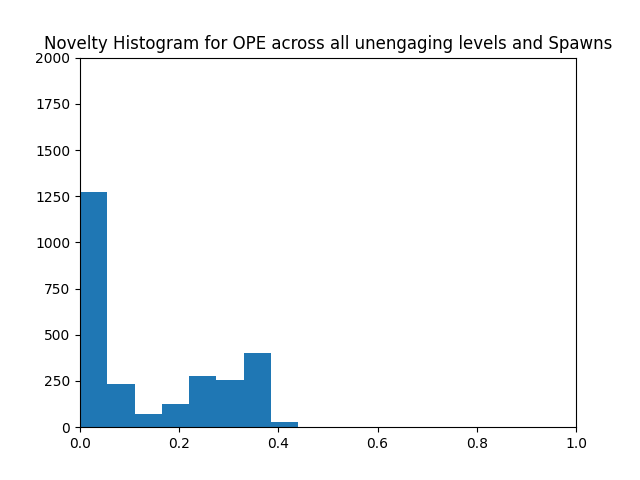}
  \caption{Openness}
\end{subfigure}\hfill 
\begin{subfigure}{.45\linewidth}
  \includegraphics[width=\linewidth, height = 2.8cm]{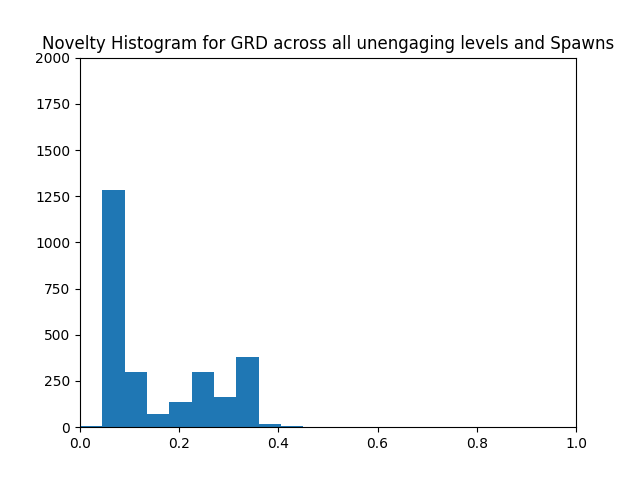}
  \caption{Group Detection}
\end{subfigure}\hfill 

\begin{subfigure}{.45\linewidth}
  \includegraphics[width=\linewidth, height = 2.8cm]{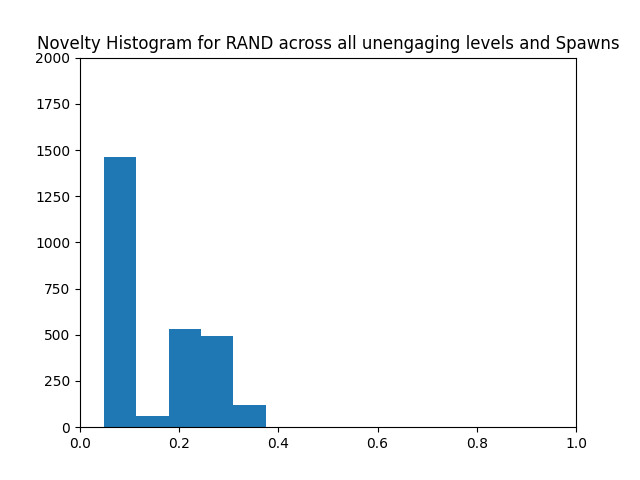}
  \caption{Random Agent}
\end{subfigure}\hfill 
\caption{Novelty Histrograms for all 3 spawns for the unengaging levels}
\label{topdown}
\end{figure}

The novelty histograms for the engaging levels compared to the unengaging levels shows significant differences for each metric. They are much higher frequencies with a low novelty, on the unengaging levels (this is particularly evident for openness, large object detection, anticipation detection and all metrics), where not much/nothing that is novel is being observed by the agent. This suggests that the engaging levels have more types of objects, spread out more evenly across the levels, so what the agent is viewing can be considered less boring. Even the random agent shows a lower overall novelty score with less peaks and dips.

\begin{figure} \label{fig:avgcoverage}
    \centering
    \includegraphics[width=0.5\linewidth]{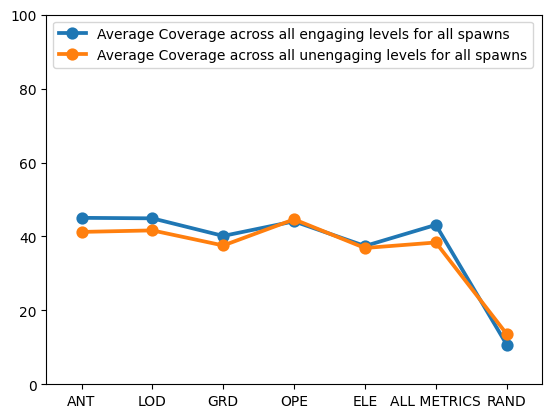}
    \caption{Average coverage for each metric in the engaging levels vs the unengaging levels}
    
\end{figure}

The average coverage does not show significant differences between all metrics at all levels. All metrics does show slightly higher average coverage on the engaging levels than the unengaging levels, as well as the random agent. This goes against our expectations. Since both the engaging and unengaging levels have roughly the same navigability, there are no physical barriers or obstacles that would prevent the agent from moving through the space. This uniformity in navigability means that the agent can traverse the entire level without being hindered, leading to similar coverage metrics across different levels, irrespective of their design or engagement factors.  

\begin{figure} \label{fig:avgentropy}
    \centering
    \includegraphics[width=0.5\linewidth]{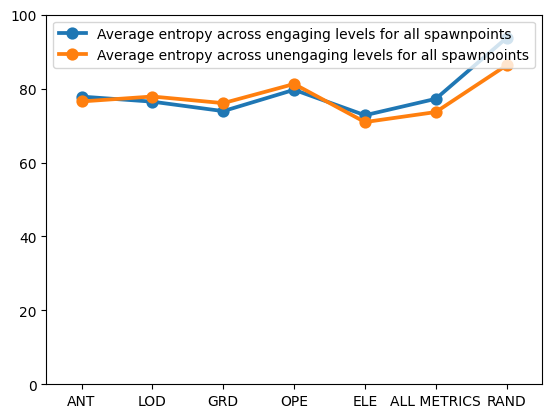}
    \caption{Average entropy for each metric in the engaging levels vs the unengaging levels}
    
\end{figure}
The average entropy does not show significant differences in entropy between the engaging and unengaging levels. Though all metrics shows slightly higher entropy in the engaging levels than the unengaging levels, which goes against our expectations. The reason why there This could be due to the agents explore the environment based on predetermined metrics and decision-making algorithms that systematically cover the space. Since the entropy measures the randomness and unpredictability of the agent's path, the systematic approach to exploration, where the agent randomly selects one of the highest scoring direction  (even if those directions score 0), can lead to similar levels of entropy regardless of the level's engaging features. The agent's behavior might inherently limit the variation in its path, leading to similar entropy values across different environments

\begin{figure}\label{fig:avginspection}
    \centering
    \includegraphics[width=0.5\linewidth]{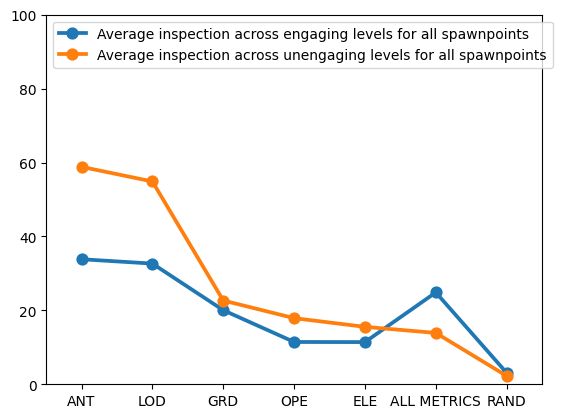}
    \caption{Average inspection for each metric in the engaging levels vs the unengaging levels}
    
\end{figure}

There are large differences between inspection, particularly for Anticipation detection, large object detection and all metrics, the unengaging levels show much higher inspection percentages (that's probably because there are less objects to investigate in the unengaging levels) However, all metrics shows much higher inspection on the engaging levels than the unengaging levels, this suggests that there was more motivation to investigate objects in the engaging levels than the unengaging.

\begin{table}[h!]
\centering
\begin{tabular}{|c|c|}
\hline
\textbf{Engaging Level} & \textbf{Fitness Score} \\ \hline
Engaging Level 1 & 0.858 \\ \hline
Engaging Level 2 & 0.916 \\ \hline
Engaging Level 3 & 0.702 \\ \hline
Engaging Level 4 & 0.728 \\ \hline
Engaging Level 5 & 0.835 \\ \hline
\end{tabular}
\caption{Fitness Scores for Engaging Levels}
\label{tab:engaging_levels}
\end{table}

\begin{table}[h!]
\centering
\begin{tabular}{|c|c|}
\hline
\textbf{Unengaging Level} & \textbf{Fitness Score} \\ \hline
Unengaging Level 1 & 0.673 \\ \hline
Unengaging Level 2 & 0.531 \\ \hline
Unengaging Level 3 & 0.645 \\ \hline
Unengaging Level 4 & 0.448 \\ \hline
Unengaging Level 5 & 0.168 \\ \hline
\end{tabular}
\caption{Fitness Scores for Unengaging Levels}
\label{tab:unengaging_levels}
\end{table}
The fitness scores for both engaging and unengaging levels, demonstrate a clear distinction in the exploratory potential of the levels generated by the two different procedural content generators. As shown in Table \ref{tab:engaging_levels}, the fitness scores for the engaging levels are consistently higher for every level, ranging from 0.703 to 0.916, with an average fitness score across all engaging levels of approximately 0.808. This suggests that the engaging levels are well-suited for exploration, offering a rich and varied environment that aligns with the exploratory behaviors of the agents.

Table \ref{tab:unengaging_levels} shows significantly lower fitness scores for the unengaging levels, with values ranging from 0.168 to 0.673 and an average fitness score of approximately 0.494. The lower scores in these levels indicate that they are less appropriate for engaging exploration, due to a lack of engaging features and/or a more repetitive and predictable structure. The other data (histograms and measurements of average inspection) also support this. The stark difference in average fitness between engaging and unengaging levels (0.808 vs. 0.494) highlights the effectiveness of our classifier in distinguishing between levels with high and low exploratory potential, further supporting the utility of exploratory agents in procedural content generation.


\section{Future Work} \label{futurework}
Future work could explore a broader range of PCG techniques beyond the current WFC constraint-based systems. For example, incorporating evolutionary algorithms with the fitness function used in this experiment to determine whether they are high-quality exploratory experiences could provide insight into how different generation strategies impact the exploratory behaviour of agents. This would allow for a more comprehensive understanding of the strengths and weaknesses of various PCG approaches. Testing a wider variety of levels, bigger or smaller in area, with a wider range of assets, could also help with this.

Also, though agent-based exploration is valuable for evaluating PCG environments, incorporating human/player feedback could enhance the understanding of how these environments support real player experiences. Conducting user studies where human players interact with the generated levels and comparing their exploration patterns and experiences with those of the agents could provide deeper insights into the alignment between agent-based metrics and actual player engagement.

\section{Conclusion} \label{conc}
In conclusion, there is evidence to suggest that our exploratory agent can distinguish between engaging and unengaging levels that have been generated procedurally. Future work should aim to test a larger sample size of generated levels and to have agents provide feedback on a generation process to produce higher quality PGC.




\begin{acknowledgments}
This work was supported by the EPSRC Centre for Doctoral Training in Intelligent Games \& Games Intelligence (iGGi) EP/S022325/1.

\end{acknowledgments}

\bibliography{Main}

\appendix

\end{document}